%% file: Vanaret.tex
\documentclass{llncs}

\usepackage{fixltx2e}
\usepackage{algorithm}
\usepackage{algpseudocode}

\MakeRobust{\Call}
\MakeRobust{\If}

\usepackage{lmodern}
\usepackage{amsmath,amsfonts,amssymb}
\usepackage{bm}
\usepackage{subfig}
\usepackage{rotating, multirow, pdflscape}
\usepackage{graphicx}
\usepackage{color}

\definecolor{amblu}{RGB}{47, 96, 107}
\definecolor{amred}{RGB}{230,0,0}
\definecolor{amgreen}{RGB}{96,107,47}

\newcommand{\inout}{\textbf{in-out}}
\newcommand{\ul}{\underline}

\title{Hybridization of Interval CP and Evolutionary Algorithms for Optimizing
Difficult Problems}
\author{C. Vanaret\inst{1} \and J-B. Gotteland\inst{2} \and N. Durand\inst{2} \and J-M.
Alliot\inst{1}}
\institute{Institut de Recherche en Informatique de Toulouse \\
2 rue Charles Camichel, 31000 Toulouse, France \\
\email{charlie.vanaret@enseeiht.fr}, \email{jean-marc.alliot@irit.fr}
\and
Ecole Nationale de l'Aviation Civile \\
7 avenue Edouard Belin, 31055 Toulouse Cedex 04, France \\
\email{gottelan@recherche.enac.fr}, \email{durand@recherche.enac.fr}}

\begin{document}

\maketitle

\begin{abstract}
The only rigorous approaches for achieving a numerical proof of optimality in global optimization
are interval-based methods that interleave branching of the search-space and pruning of the
subdomains that cannot contain an optimal solution. State-of-the-art solvers generally
\textit{integrate} local optimization algorithms to compute a good upper bound of the global minimum
over each subspace. In this document, we propose a \textit{cooperative} framework in which interval
methods cooperate with evolutionary algorithms. The latter are stochastic algorithms in which a
population of candidate solutions iteratively evolves in the search-space to reach satisfactory
solutions.

Within our cooperative solver Charibde, the evolutionary algorithm and the interval-based
algorithm run in parallel and exchange bounds, solutions and search-space in an advanced manner via
message passing. A comparison of Charibde with state-of-the-art interval-based solvers (GlobSol,
IBBA, Ibex) and NLP solvers (Couenne, BARON) on a benchmark of difficult COCONUT problems shows that
Charibde is highly competitive against non-rigorous solvers and converges faster than rigorous
solvers by an order of magnitude.
\end{abstract}

\section{Motivation}
We consider $n$-dimensional continuous constrained optimization problems over a
hyperrectangular domain $\bm{D} = D_1 \times \ldots \times D_n$:
\begin{equation}
\begin{aligned}
(\mathcal{P}) \quad & \min_{\bm{x} \in \bm{D} \subset \mathbb{R}^n} 	& f(\bm{x}) & \\
					& s.t.	& g_i(\bm{x}) \le 0, & \quad i \in \{1, \ldots, m\} \\
					& 		& h_j(\bm{x}) = 0, & \quad j \in \{1, \ldots, p\} \\
\end{aligned}
\end{equation}

When $f$, $g_i$ and $h_j$ are non-convex, the problem may have multiple local minima. Such difficult
problems are generally solved using generic exhaustive branch and bound (BB) methods. The objective
function and constraints are bounded on disjoint subspaces using enclosure methods. By keeping
track of the best known upper bound $\tilde{f}$ of the global minimum $f^*$, subspaces that cannot 
contain a global
minimizer are discarded (pruned).

Several authors proposed hybrid approaches in which a BB algorithm cooperates with another technique
to enhance the pruning of the search-space. Hybrid algorithms may be classified into two
categories~\cite{Puchinger2005Combining}: \textit{integrative} approaches, in which one of the two
methods replaces a particular operator of the other method, and \textit{cooperative} methods, in
which the methods are independent and are run sequentially or in parallel. Previous works include

\begin{itemize}
\item integrative approaches:
\cite{Zhang2007New} integrates a stochastic genetic algorithm (GA) within an interval BB. The GA
provides the direction along which a box is partitioned, and an individual is generated within
each subbox.
At each generation, the best evaluation updates the best known upper bound of the global minimum.
In \cite{Cotta2003Embedding}, the crossover operator is replaced by a BB that determines the best
individual among the offspring.

\item cooperative approaches:
\cite{Sotiropoulos1997New} sequentially combines an interval BB and a GA. The interval BB generates
a list $\mathcal{L}$ of remaining small boxes. The GA's population is initialized by generating a
single individual within each box of $\mathcal{L}$.
\cite{Gallardo2007Hybridization} (BB and memetic algorithm) and \cite{Blum2011Hybrid} (beam search
and memetic algorithm) describe similar parallel strategies: the BB identifies promising regions
that are then explored by the metaheuristic.
% \cite{Cotta1995Hybridizing} évoque la possibilité de combiner en parallèle un BB et un
% GA sur le problème du voyageur de commerce. L'GA fournit au BB un majorant du
% minimum global, afin d'éliminer les sous-problèmes sous-optimaux. Le BB
% injecte dans la population de l'GA les circuits qu'il considère prometteurs. Les auteurs
% critiquent néanmoins cette approche : échanger des informations entre des processus
% s'exécutant à des vitesses différentes est une tâche difficile. En particulier, injecter des
% solutions prometteuses en début de convergence de l'GA risque de faire apparaître des ``
% superindividus '' et de réduire la diversité de la population. Deux approches alternatives sont
% considérées : une hybridation intégrative (remplacer le croisement de l'GA par une
% recherche arborescente) ou un schéma maître-esclaves (composé d'un BB et de $m$ GA en
% parallèle).
\cite{Alliot2012Finding} hybridizes a GA and an interval BB. The two independent algorithms exchange
upper bounds and solutions through shared memory. New optimal results are presented for the rotated
Michalewicz ($n = 12$) and Griewank functions ($n = 6$).
\end{itemize}

In this communication, we build upon the cooperative scheme of~\cite{Alliot2012Finding}. The
efficiency and reliability of their solver remain very limited; it is not competitive against
state-of-the-art solvers. Their interval techniques are naive and may lose solutions,
% (boxes whose
% width is lower than a given treshold are discarded),
while the GA may send
evaluations subject to roundoff errors. We propose to hybridize a stochastic differential evolution
algorithm (close to a GA), described in Section \ref{sec:de}, and a deterministic
interval branch and contract algorithm, described in Section \ref{sec:ibc}. Our hybrid solver
Charibde is presented in Section \ref{sec:charibde}. Experimental results (Section
\ref{sec:results}) show that Charibde is highly competitive against state-of-the-art solvers.

\section{Differential Evolution}\label{sec:de}
Differential evolution (DE)~\cite{StornPrice1997} is among the simplest and most efficient
metaheuristics for continuous problems. It combines the coordinates of existing individuals
(candidate solutions) with a
given probability to generate new individuals. Initially devised for
continuous unconstrained problems, DE was extended to mixed problems and constrained
problems~\cite{PriceStornLampinen2006}.
% Noteworthy results of DE include training of neural networks~\cite{Slowik2008Training},
% aerodynamic design~\cite{Rogalsky2000Differential}, multiple-criteria decision analysis,
% polynomial approximation and task scheduling.

% \begin{algorithm}[h!]
% \caption{Differential Evolution}
% \label{alg:ed}
% \begin{algorithmic}[]
% \Function{DifferentialEvolution}{$f$: objective function, $\mathcal{C}$: system of constraints,
% $\bm{D}$: search-space, $\mathit{NP}$: size of population, $W$: amplitude factor, $\mathit{CR}$:
% crossover rate}
% \State $P \gets$ initial population, randomly chosen in $\bm{D}$
% \Repeat
% 	\State $P' \gets \varnothing$
% 	\Comment temporary population
% 	\For{$\bm{x} \in P$}
% 		\State $(\bm{u}, \bm{v}, \bm{w}) \gets$ \Call{ChooseParents}{$\bm{x}$, $P$}
% 		\State $\bm{y} \gets$ \Call{Crossover}{$\bm{x}$, $\bm{u}$, $\bm{v}$, $\bm{w}$, $W$,
% $\mathit{CR}$}
% 		\Comment new individual
% 		\If{$\bm{y}$ is better than $\bm{x}$}
% 			\State $P' \gets P' \cup \{\bm{y}\}$
% 			\Comment $\bm{y}$ replaces $\bm{x}$
% 		\Else 
% 			\State $P' \gets P' \cup \{\bm{x}\}$
% 			\Comment $\bm{x}$ is kept
% 		\EndIf
% 	\EndFor
% 	\State $P \gets P'$
% \Until{termination criterion is met} \\
% \Return best individual of $P$
% \EndFunction
% \end{algorithmic}
% \end{algorithm}

Let $\mathit{NP}$ denote the size of the population, $W > 0$ the amplitude factor and $\mathit{CR}
\in [0, 1]$ the crossover rate. At each generation (iteration), $\mathit{NP}$ new individuals are
generated: for each individual $\bm{x} = (x_1, \ldots, x_n)$, three other individuals $\bm{u} =
(u_1, \ldots, u_n)$ (called \textit{base individual}), $\bm{v} = (v_1, \ldots, v_n)$ and $\bm{w} =
(w_1, \ldots, w_n)$, all different and different from $\bm{x}$, are randomly picked in
the population. The coordinates $y_i$ of the new individual $\bm{y} = (y_1, \ldots, y_n)$ are
computed according to

\begin{equation}
y_i =
\begin{cases}
u_i + W \times (v_i - w_i) 	& \text{if } i = R \text{ or } r_i < \mathit{CR} \\
x_i							& \text{otherwise}
\end{cases}
\end{equation}
where $r_i$ is picked in $[0, 1]$ with uniform probability. The index $R$, picked in $\{1, \ldots,
n\}$ with uniform probability for each $\bm{x}$, ensures that at least a coordinate of $\bm{y}$
differs from that of $\bm{x}$. $\bm{y}$ replaces $\bm{x}$ in the population if it is ``better''
than $\bm{x}$ (e.g. in unconstrained optimization, $\bm{y}$ is better than $\bm{x}$ if it improves 
the
objective function). 

\begin{figure}[h!]
\centering
\def\svgwidth{0.5\columnwidth}
\small
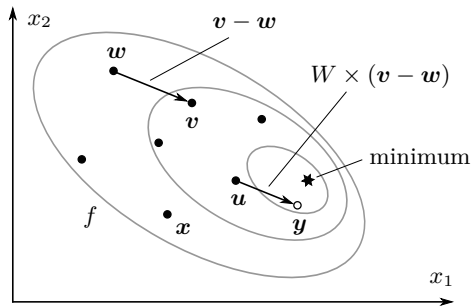
\caption{Crossover of the differential evolution}
\label{fig:de}
\end{figure}

Figure \ref{fig:de} depicts a two-dimensional crossover between individuals $\bm{x}$, $\bm{u}$
(base individual), $\bm{v}$ and $\bm{w}$. The contour lines of the objective function are shown in
grey. The difference $\bm{v}-\bm{w}$, scaled by $W$, yields the direction (an approximation of the
direction opposite the gradient) along which $\bm{u}$ is translated to yield $\bm{y}$.

\section{Reliable Computations}\label{sec:ibc}
\textit{Reliable} (or rigorous) methods provide bounds on the global minimum, even in the presence
of roundoff errors. The only reliable approaches for achieving a numerical proof of optimality in
global optimization are interval-based methods that interleave branching of the search-space and
pruning of the subdomains that cannot contain an optimal solution.

Section \ref{sec:ia} introduces interval arithmetic, an extension of real
arithmetic. Reliable global optimization is detailed in Section \ref{sec:optimization}, and
interval contractors are mentioned in Section \ref{sec:contractors}.

\subsection{Interval Arithmetic}\label{sec:ia}

An \textit{interval} $X$ with floating-point bounds defines the set $\{ x \in \mathbb{R} ~|~
\underline{X} \le x \le \overline{X} \}$. $\mathbb{IR}$ denotes the set of all intervals. The
\textit{width} of $X$ is $w(X) = \overline{X} - \underline{X}$. $m(X) = \frac{\underline{X} +
\overline{X}}{2}$ is the \textit{midpoint} of $X$. A \textit{box} $\bm{X}$ is a Cartesian product
of intervals. The width of a box is the maximum width of its components. The \textit{convex hull}
$\square(X, Y)$ of $X$ and $Y$ is the smallest interval enclosing $X$ and $Y$.

Interval arithmetic~\cite{Moore1966} extends real arithmetic to intervals. Interval arithmetic
implemented on a machine must be \textit{rounded outward} (the left bound is rounded toward
$-\infty$, the right bound toward $+\infty$) to guarantee conservative properties.
The interval counterparts of binary operations and elementary functions produce the smallest
interval containing the image. Thanks to the conservative properties of interval arithmetic, we
define interval extensions (Definition \ref{def:interval-ext}) of functions that may be expressed 
as a
finite composition of elementary functions.

\begin{definition}[Interval extension]\label{def:interval-ext}
Let $f : \mathbb{R}^n \rightarrow \mathbb{R}$. $F : \mathbb{IR}^n \rightarrow \mathbb{IR}$ is an
\emph{interval extension} (or inclusion function) of $f$ iff
\begin{equation}
\begin{aligned}
\forall \bm{X} \in \mathbb{IR}^n, & \quad f(\bm{X}) := \{ f(\bm{x}) ~|~ \bm{x} \in \bm{X} \} \subset
F(\bm{X}) \\
\forall \bm{X} \in \mathbb{IR}^n, \forall \bm{Y} \in \mathbb{IR}^n, & \quad \bm{X} \subset \bm{Y}
\Rightarrow F(\bm{X}) \subset F(\bm{Y})
\end{aligned}
\end{equation}
\end{definition}

Interval extensions with various sharpnesses may be defined (Example \ref{ex:interval-ex}). The
\textit{natural interval extension} $F_N$ replaces the variables with their domains and
the elementary functions with their interval counterparts. The \textit{Taylor interval extension}
$F_T$ is based on the Taylor expansion at point $\bm{c} \in \bm{X}$.
% \begin{equation}
% F_T(\bm{X}, \bm{c}) = F(\bm{c}) + \sum_{i=1}^n \frac{\partial F}{\partial x_i}(\bm{X}) \cdot
% (X_i-c_i)
% \end{equation}

\begin{example}[Interval extensions]\label{ex:interval-ex}
Let $f(x) = x^2-x$, $X = [-2, 0.5]$ and $c = -1 \in X$. The exact range is $f(X) = [-0.25, 6]$. Then
\begin{itemize}
\item $F_N(X) = X^2 - X = [-2, 0.5]^2 - [-2, 0.5] = [0, 4] - [-2, 0.5] = [-0.5, 6]$;
\item $F_T(X, c) = 2 + (2X-1)(X+1) = 2 + [-5, 0][-1, 1.5] = [-5.5, 7]$.
\end{itemize}
\end{example}

Example \ref{ex:interval-ex} shows that interval arithmetic often overestimates the range of a
real-valued function. This is due to the \textit{dependency problem}, an inherent behavior of
interval arithmetic. Dependency decorrelates multiple occurrences of the same variable in an
analytical expression (Example \ref{ex:dependency}).

\begin{example}[Dependency]\label{ex:dependency}
Let $X = [-5, 5]$. Then
\begin{equation}
\begin{aligned}
 X - X = [-10, 10] 	& = \{ x_1 - x_2 ~|~ x_1 \in X, x_2 \in X \} \\
					& \supset \{ x - x ~|~ x \in X \} = \{0\}
\end{aligned}
\end{equation}
\end{example}

Interval extensions $(F_N, F_T)$ have different convergence orders, that is the overestimation
decreases at different speeds with the width of the interval.

% L'évaluation d'expressions syntaxiquement équivalentes en arithmétique réelle peut fournir des
% encadrements plus ou moins précis en AI.

\subsection{Global Optimization}\label{sec:optimization}

Interval arithmetic computes a rigorous enclosure of the range of a function over a box. The first
branch and bound algorithms for continuous optimization based on interval arithmetic were devised in
the 1970s~\cite{Moore1976Computing}\cite{Skelboe1974Computation}, then refined during the following
years~\cite{Hansen1992}: the search-space is partitioned into subboxes. The objective function and
constraints are evaluated on each subbox using interval arithmetic. The subspaces that cannot
contain a global minimizer are discarded and are not further explored.  The algorithm terminates
when $\tilde{f}-f^* < \varepsilon$.

To overcome the pessimistic enclosures of interval arithmetic, interval branch and bound algorithms
have recently been endowed with filtering algorithms (Section \ref{sec:contractors}) that
narrow the bounds of the boxes without loss of solutions. Stemming from the Interval Analysis and
Interval Constraint Programming communities, filtering (or contraction) algorithms discard values
from the domains by enforcing local (each constraint individually) or global (all constraints
simultaneously) consistencies. The resulting methods, called interval branch and contract (IBC)
algorithms, interleave steps of contraction and steps of bisection.

\subsection{Interval Contractors}\label{sec:contractors}

State-of-the-art contractors (contraction algorithms) include HC4~\cite{Benhamou1999},
Box~\cite{VanHentenryck1997Numerica}, Mohc~\cite{Araya2010}, 3B~\cite{Lhomme1993Consistency},
CID~\cite{Trombettoni2007Constructive} and X-Newton~\cite{Araya2012Contractor}. Only HC4 and
X-Newton are used in this communication.

HC4Revise is a two-phase algorithm that exploits the syntax tree of a constraint to contract each
occurrence of the variables. The first phase (evaluation) evaluates each node (elementary function)
using interval arithmetic. The second phase (propagation) uses projection functions to inverse each
elementary function.
% HC4Revise is optimal when variables have a single occurrence in the expression
% of the constraint (when the constraint and projection functions are continuous).
HC4Revise is generally invoked as the revise procedure (subcontractor) of HC4, an AC3-like 
propagation
loop.

X-Newton computes an outer linear relaxation of the objective function and the constraints, then
computes a lower bound of the initial problem using LP techniques (e.g. the simplex algorithm).
$2n$ additional calls may contract the domains of the variables.

\section{Charibde: a Cooperative Approach}\label{sec:charibde}

\subsection{Hybridization of Stochastic and Deterministic Techniques}

Our hybrid algorithm \textit{Charibde} (Cooperative Hybrid Algorithm using Reliable Interval-Based
methods and Differential Evolution), written in OCaml~\cite{Leroy2010OCaml}, combines a stochastic
DE and a deterministic IBC for non-convex constrained optimization. Although it embeds a stochastic
component, Charibde is a \textit{fully rigorous} solver.

\subsubsection{Previous Work}

Preliminary results of a basic version of Charibde were published in
2013~\cite{Vanaret2013Preventing}
on classical multimodal problems (7 bound-constrained and 4 inequality-constrained problems) widely
used in the Evolutionary Computation community. We showed that Charibde benefited from the start
of convergence of the DE algorithm, and completed the proof of convergence faster than a standard
IBC algorithm. We provided new optimal results for 3 problems (Rana, Eggholder and
Michalewicz).

\subsubsection{Contributions}

In this communication, we present \textit{two contributions}:
\begin{enumerate}
\item we devised a new cooperative exploration strategy MaxDist that
	\begin{itemize}
	\item selects boxes to be explored in a novel manner;
	\item periodically reduces DE's domain;
	\item restarts the population within the new (smaller) domain.
	\end{itemize}
	An example illustrates the evolution of the domain without loss of solutions;
\item we assess the performance of Charibde against state-of-the-art rigorous (GlobSol, IBBA,
Ibex) and non-rigorous (Couenne, BARON) solvers on a benchmark of difficult problems.
\end{enumerate}

\subsubsection{Cooperative Scheme}
\begin{figure}[htbp]
\centering
\def\svgwidth{0.78\columnwidth}
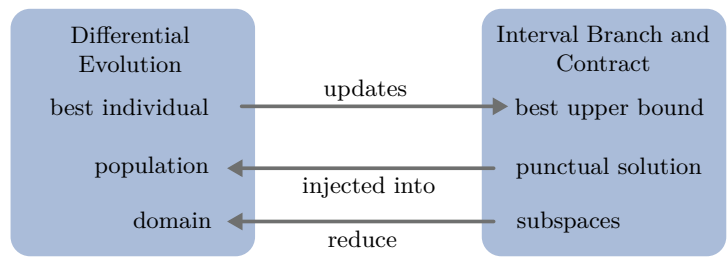
\caption{Cooperative scheme of Charibde}
\label{fig:charibde}
\end{figure}

Two independent parallel processes exchange bounds, solutions and search-space via MPI message
passing (Figure \ref{fig:charibde}).

% \clearpage

The cooperation boils down to three main steps:
\begin{enumerate}
\item whenever the DE improves its best evaluation, the best individual and its evaluation are sent
to the IBC to update the best known upper bound $\tilde{f}$;
\item whenever the IBC finds a better punctual solution (e.g. the center of a box), it is injected
into DE's population;
\item the exploration strategy MaxDist periodically reduces the search-space of DE, then
regenerates the population in the new search-space.
\end{enumerate}

Sections \ref{sec:charibde-de} and \ref{sec:charibde-ibc} detail the particular implementations
of the DE (Algorithm \ref{alg:charibde-de}) and the IBC (Algorithm
\ref{alg:charibde-ibc}) within Charibde.

\begin{algorithm}[h!]
\caption{Charibde: Differential Evolution}
\label{alg:charibde-de}
\footnotesize
\begin{algorithmic}[]
\Function{DifferentialEvolution}{$f$: objective function, $\mathcal{C}$: system of constraints,
$\bm{D}$: search-space, $\mathit{NP}$: size of population, $W$: amplitude factor, $\mathit{CR}$:
crossover rate}
\State $P \gets$ initial population, randomy generated in $\bm{D}$
\textcolor{amred}{\State $\tilde{f} \gets +\infty$}
\Repeat
	\State \textcolor{amblu}{$(\bm{x}, f_{\bm{x}}) \gets$ MPI\_ReceiveIBC()}
	\State \textcolor{amblu}{Insert $\bm{x}$ into $P$}
	\State \textcolor{amblu}{$\tilde{f} \gets f_{\bm{x}}$}
	\State Generate temporary population $P'$ by crossover
	\State $P \gets P'$
	\State $(\bm{x}_{best}, f_{best}) \gets$ \Call{BestIndividual}{$P$}
	\If{$f_{best} < \tilde{f}$}
		\State \textcolor{amred}{$\tilde{f} \gets f_{best}$}
		\State \textcolor{amred}{MPI\_SendIBC$(\bm{x}_{best}, f_{best})$}
	\EndIf
\Until{termination criterion is met}
\State \Return best individual of $P$
\EndFunction
\end{algorithmic}
\end{algorithm}

\begin{algorithm}[h!]
\caption{Charibde: Interval Branch and Contract}
\label{alg:charibde-ibc}
\footnotesize
\begin{algorithmic}
\Function{IntervalBranchAndContract}{$F$: objective function, $\mathcal{C}$: system of constraints,
$\bm{D}$: search-space, $\varepsilon$: precision}
\State $\tilde{f} \gets +\infty$
\Comment best known upper bound
\State $\mathcal{Q} \gets \{ \bm{D} \}$
\Comment priority queue
\While{$\mathcal{Q} \neq \varnothing$}
	\State \textcolor{amred}{$(\bm{x}_{DE}, f_{DE}) \gets$ MPI\_ReceiveDE$()$}
	\State \textcolor{amred}{$\tilde{f} \gets \min(\tilde{f}, f_{DE})$}
	\State Extract a box $\bm{X}$ from $\mathcal{Q}$
	\State Contract $\bm{X}$ w.r.t. constraints
	\Comment Algorithm \ref{algo:charibde-contractor}
	\If{$\bm{X}$ cannot be discarded}
		\If{$\overline{F(m(\bm{X}))} < \tilde{f}$}
			\Comment midpoint test
			\State $\tilde{f} \gets \overline{F(m(\bm{X}))}$
			\Comment update best upper bound
			\State \textcolor{amblu}{MPI\_SendDE$(m(\bm{X}), \overline{F(m(\bm{X}))})$}
		\EndIf
		\State Split $\bm{X}$ into $\{\bm{X}_1, \bm{X}_2\}$
		\State Insert $\{\bm{X}_1, \bm{X}_2\}$ into $\mathcal{Q}$
	\EndIf
\EndWhile
\State \Return $(\tilde{f}, \tilde{\bm{x}})$
\EndFunction
\end{algorithmic}
\end{algorithm}

%%%%%%%%%%%%%%%%%%%%%%%%%%%%
% Evolution Diff\'erentielle %
%%%%%%%%%%%%%%%%%%%%%%%%%%%%
\subsection{Differential Evolution}\label{sec:charibde-de}
Population-based metaheuristics, in particular DE, are endowed with mechanisms that help escape
local minima. They are quite naturally recommended to solve difficult multimodal problems for which
traditional methods struggle to converge. They are also capable of generating feasible solutions
without any a priori knowledge of the topology. DE has proven greatly beneficial for improving the
best known upper bound $\tilde{f}$, a task for which standard branch and bound algorithms are not
intrinsically intended.

\subsubsection{Base Individual}
In the standard DE strategy, all the current individuals have the same probability to be selected
as the base individual $\bm{u}$. We opted for an alternative strategy~\cite{PriceStornLampinen2006}
that guarantees that all individuals of the population play this role once and only once at each
generation: the index of the base individual is obtained by summing the index of the individual
$\bm{x}$ and an offset in $\{1, \ldots, \mathit{NP}-1\}$, drawn with uniform probability.

\subsubsection{Bound Constraints}
When a coordinate $y_i$ of $\bm{y}$ (computed during the crossover) exceeds the bounds of the
component $D_i$ of the domain $\bm{D}$,
the bounce-back method~\cite{PriceStornLampinen2006} replaces $y_i$ with a valid coordinate $y_i'$
that lies between the base coordinate $u_i$ and the violated bound:
\begin{equation}
y_i' =
\begin{cases}
u_i + \omega(\overline{D_i} - u_i) 	& \text{if } y_i > \overline{D_i} \\
u_i + \omega(\underline{D_i} - u_i) 	& \text{if } y_i < \underline{D_i}
\end{cases}
\end{equation}
where $\omega$ is drawn in $[0, 1]$ with uniform probability.

\subsubsection{Constraint Handling}
The extension of evolutionary algorithms to constrained optimization has been addressed by numerous
authors. We implemented the direct constraint handling~\cite{PriceStornLampinen2006} that assigns
to each individual a vector of evaluations (objective function and constraints), and selects the new
individual $\bm{y}$ (see Section \ref{sec:de}) based upon simple rules:
\begin{itemize}
\item $\bm{x}$ and $\bm{y}$ are feasible and $\bm{y}$ has a lower or equal objective value
than $\bm{x}$;
\item $\bm{y}$ is feasible and $\bm{x}$ is not;
\item $\bm{x}$ and $\bm{y}$ are infeasible, and $\bm{y}$ does not violate any constraint
more than $\bm{x}$.
\end{itemize}
% We slightly adapted the direct constraint handling by introducing an additional comparison on the
% number and the magnitudes of the violated constraints. Let us denote for an individual $\bm{x}$:
% \begin{itemize}
% \item $f_{\bm{x}}$ the objective value of $\bm{x}$;
% \item $n_{\bm{x}}$ the number of constraints ($\in \{1, \ldots, m\}$) violated by $\bm{x}$;
% \item $s_{\bm{x}} \eqdef \sum_{i=1}^m \max(0, g_i(\bm{x}))$ the sum of magnitudes of the
% constraints violated by $\bm{x}$.
% \end{itemize}
% We assign to $\bm{x}$ either one of the statuses:
% \begin{itemize}
% \item $\mathit{Feasible}(f_{\bm{x}})$ when $\bm{x}$ is a feasible individual;
% \item $\mathit{Infeasible}(n_{\bm{x}}, s_{\bm{x}})$ when $\bm{x}$ is not feasible. If at least one
% of the constraints is violated, the objective function is not evaluated.
% \end{itemize}
% 
% Rule 3 of~\cite{PriceStornLampinen2006} is then replaced by:
% \begin{enumerate}
% \setcounter{enumi}{2}
% \item both $\bm{x}$ and $\bm{y}$ are infeasible, and $n_{\bm{y}} < n_{\bm{x}}$ or ($n_{\bm{y}} =
% n_{\bm{x}}$ and $s_{\bm{y}} < s_{\bm{x}}$).
% \end{enumerate}

\subsubsection{Rigorous Feasibility}
Numerous NLP solvers tolerate a slight violation (relaxation) of the inequality constraints (e.g.
$g \le 10^{-6}$ instead of $g \le 0$). The evaluation of a ``pseudo-feasible'' solution $\bm{x}$
(that satisfies such relaxed constraints) is not a rigorous upper bound of the global minimum;
roundoff errors may even lead to absurd conclusions: $f(\bm{x})$ may be lower than the global
minimum, and (or) $\bm{x}$ may be very distant from actual feasible solutions in the search-space.

To ensure that an individual $\bm{x}$ is numerically feasible (i.e. that the evaluations of the
constraints are nonpositive), we evaluate the constraints $g_i$ using interval arithmetic. $\bm{x}$
is considered as feasible when the interval evaluations $G_i(\bm{x})$ are nonpositive,
that is $\forall i \in \{1, \ldots, m\}, \overline{G_i(\bm{x})} \le 0$.

\subsubsection{Rigorous Objective Function}
When $\bm{x}$ is a feasible point, the evaluation $f(\bm{x})$ may be subject to roundoff errors;
the only reliable upper bound of the global minimum available is $\overline{F(\bm{x})}$ (the right
bound of the interval evaluation). However, evaluating the individuals using only interval
arithmetic is much costlier than cheap floating-point arithmetic.

An efficient in-between solution consists in systematically computing the floating-point
evaluations $f(\bm{x})$, and computing the interval evaluation $F(\bm{x})$ only when the best known
approximate evaluation is improved. $\overline{F(\bm{x})}$ is then compared to the best
known reliable upper bound $\tilde{f}$: if $\tilde{f}$ is improved, $\overline{F(\bm{x})}$ is sent
to the IBC. This implementation greatly reduces the cost of evaluations, while ensuring that all
the values sent to the IBC are rigorous.

%%%%%%%%%%%%%%%%%%%%%%%
% Branch and Contract %
%%%%%%%%%%%%%%%%%%%%%%%
\subsection{Interval Branch and Contract}\label{sec:charibde-ibc}

\textit{Branching} aims at refining the computation of lower bounds of the functions using interval
arithmetic. Two strategies may be found in the early literature:
\begin{itemize}
\item the variable with the largest domain is bisected;
\item the variables are bisected one after the other in a round-robin scheme.
\end{itemize}
More recently, the Smear heuristic~\cite{Csendes1997Subdivision} has emerged as a competitive
alternative to the two standard strategies. The variable $x_i$ for which the interval quantity
$\frac{\partial F}{\partial x_i}(\bm{X}) (X_i - x_i)$ is the largest is bisected.
% It amounts to selecting the variable for which the Taylor form varies the most.

Charibde's \textit{main contractor} is detailed in Algorithm \ref{algo:charibde-contractor}. We
exploit the contracted nodes of HC4Revise to compute partial derivatives via automatic
differentiation~\cite{Schichl2005Interval}. HC4Revise is a revise procedure within a quasi-fixed
point algorithm with tolerance $\eta \in [0, 1]$: the propagation loop stops when the box $\bm{X}$
is not sufficiently contracted, i.e. when the size of $\bm{X}$ becomes larger than a fraction $\eta
w_0$ of the initial size $w_0$. Most contractors include an evaluation phase that yields a lower
bound of the problem on the current box. Charibde thus computes several lower bounds (natural,
Taylor, LP) as long as the box is not discarded.
Charibde calls ocaml-glpk~\cite{OCamlGlpk2004}, an OCaml binding for GLPK (GNU Linear Programming
Kit). Since the solution of the linear program is computed using floating-point arithmetic, it may
be subject to roundoff errors. A cheap postprocessing step~\cite{Neumaier2004Safe} computes a
rigorous bound on the optimal solution of the linear program, thus providing a rigorous lower
bound of the initial problem.

\begin{algorithm}[htb]
\caption{Charibde: contractor for constrained optimization}
\label{algo:charibde-contractor}
\begin{algorithmic}
\Function{Contraction}{\inout{} $\bm{X}$: box, $F$: objective function, \inout{} $\tilde{f}$:
best upper bound, \inout{} $\mathcal{C}$: system of constraints}
\State $lb \gets -\infty$
\Comment lower bound
\Repeat
	\State $w_0 \gets w(\bm{X})$
	\Comment{initial size}
	\State $F_{\bm{X}} \gets$ \Call{HC4Revise}{$F(\bm{X}) \le \tilde{f}$}
	\Comment evaluation of $f$/contraction
	\State $lb \gets \underline{F_{\bm{X}}}$
	\Comment lower bound by natural form
	\State $\bm{G} \gets \nabla F(\bm{X})$
	\Comment gradient by AD
	\State $lb \gets \max(lb,$ \Call{SecondOrder}{$\bm{X}, F, \tilde{f}, \bm{G}$}$)$
	\Comment{second-order form}
% 	\EndIf
% 	\If{use monotonicity}
% 		\State $lb \gets$ \Call{Monotonicity}{$\bm{X}, F, lb, \tilde{f}, \mathcal{C}$}
% 	\EndIf
	\State $\mathcal{C} \gets$ \Call{HC4}{$\bm{X}, \mathcal{C}, \eta$}
	\Comment quasi-fixed point with tolerance $\eta$
	\If{use linearization}
		\State $lb \gets \max(lb, $ \Call{Linearization}{$\bm{X}, F, \tilde{f}, \bm{G},
\mathcal{C}$}$)$
		\Comment{simplex or X-Newton}
	\EndIf
\Until{$\bm{X} = \varnothing$ or $w(\bm{X}) > \eta w_0$}
\State \Return $lb$
\EndFunction
\end{algorithmic}
\end{algorithm}

When the problem is subject to \textit{equality constraints} $h_j$ $(j \in \{1, \ldots, p\})$, IBBA
\cite{Ninin2010Reliable}, Ibex \cite{Trombettoni2011Inner} and Charibde handle a relaxed problem
where each equality constraint $h_j(\bm{x}) = 0$ $(j \in \{1, \ldots, p\})$ is replaced by two
inequalities: 
\begin{equation}
{-}\varepsilon_= \le h_j(\bm{x}) \le \varepsilon_=
\end{equation}
$\varepsilon_=$ may be chosen arbitrarily small.

\subsection{MaxDist: a New Exploration Strategy}\label{sec:heuristique-maxdist}
The boxes that cannot be discarded are stored in a priority queue $\mathcal{Q}$ to be processed at a
later stage. The order in which the boxes are extracted determines the exploration strategy of the
search-space (``best-first'', ``largest first'', ``depth-first'').
Numerical tests suggest that
\begin{itemize}
\item the ``best-first'' strategy is rarely relevant because of the overestimated range (due to the
dependency problem);
\item the ``largest first'' strategy does not give advantage to promising regions;
\item the ``depth-first'' strategy tends to quickly explore the neighborhood of local minima, but
struggles to escape from them.
\end{itemize}

We propose a new exploration strategy called MaxDist. It consists in extracting from $\mathcal{Q}$
the box that is \textit{the farthest from the current solution} $\tilde{\bm{x}}$. The underlying
ideas are to
\begin{itemize}
\item explore the neighborhood of the global minimizer (a tedious task when the objective function
is flat in this neighborhood) only when the best possible upper bound is available;
\item explore regions of the search-space that are hardly accessible by the DE algorithm.
\end{itemize}
The distance between a point $\bm{x}$ and a box $\bm{X}$ is the sum of the distances between each
coordinate $x_i$ and the closest bound of $X_i$.
Note that MaxDist is an adaptive heuristic: whenever the best known solution $\tilde{\bm{x}}$ is
updated, $\mathcal{Q}$ is reordered according to the new priorities of the boxes.

Preliminary results (not presented here) suggest that MaxDist is competitive with standard
strategies. However, the most interesting observation lies in the behavior of $\mathcal{Q}$: when
using MaxDist, the maximum size of $\mathcal{Q}$ (the maximum number of boxes simultaneously stored
in $\mathcal{Q}$) remains remarkably low (a few dozens compared to several thousands for standard
strategies). This offers promising perspectives for the cooperation between DE and IBC: the
remaining boxes of the IBC may be exploited in the DE to avoid exploring regions of the
search-space that have already been proven infeasible or suboptimal.

\begin{figure}[h!]
	\centering
	\subfloat[Initial domain]{\label{fig:initial-domain}
		\includegraphics[width=0.45\columnwidth]{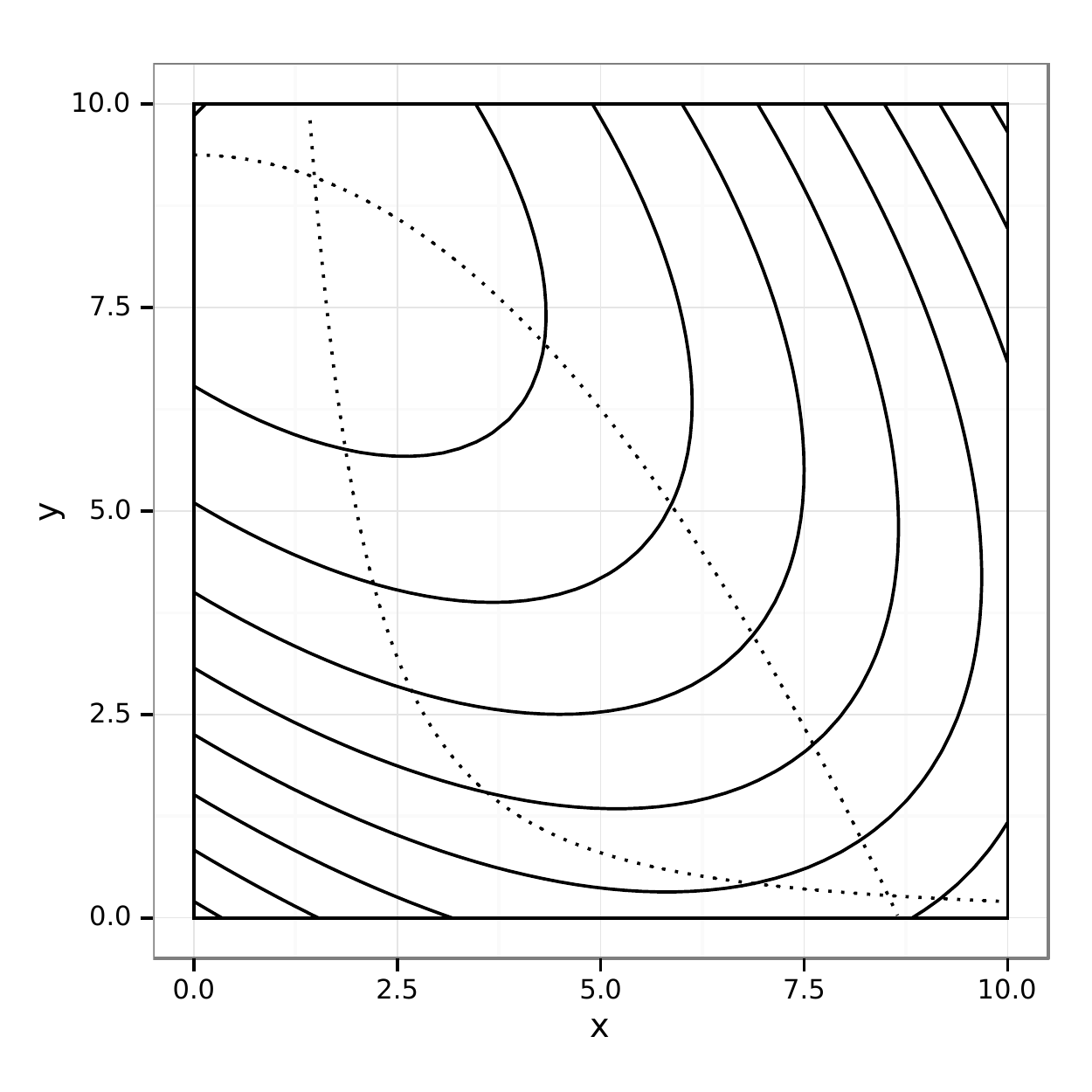}}
	\subfloat[Contracted initial domain]{\label{fig:contracted-domain}
		\includegraphics[width=0.45\columnwidth]{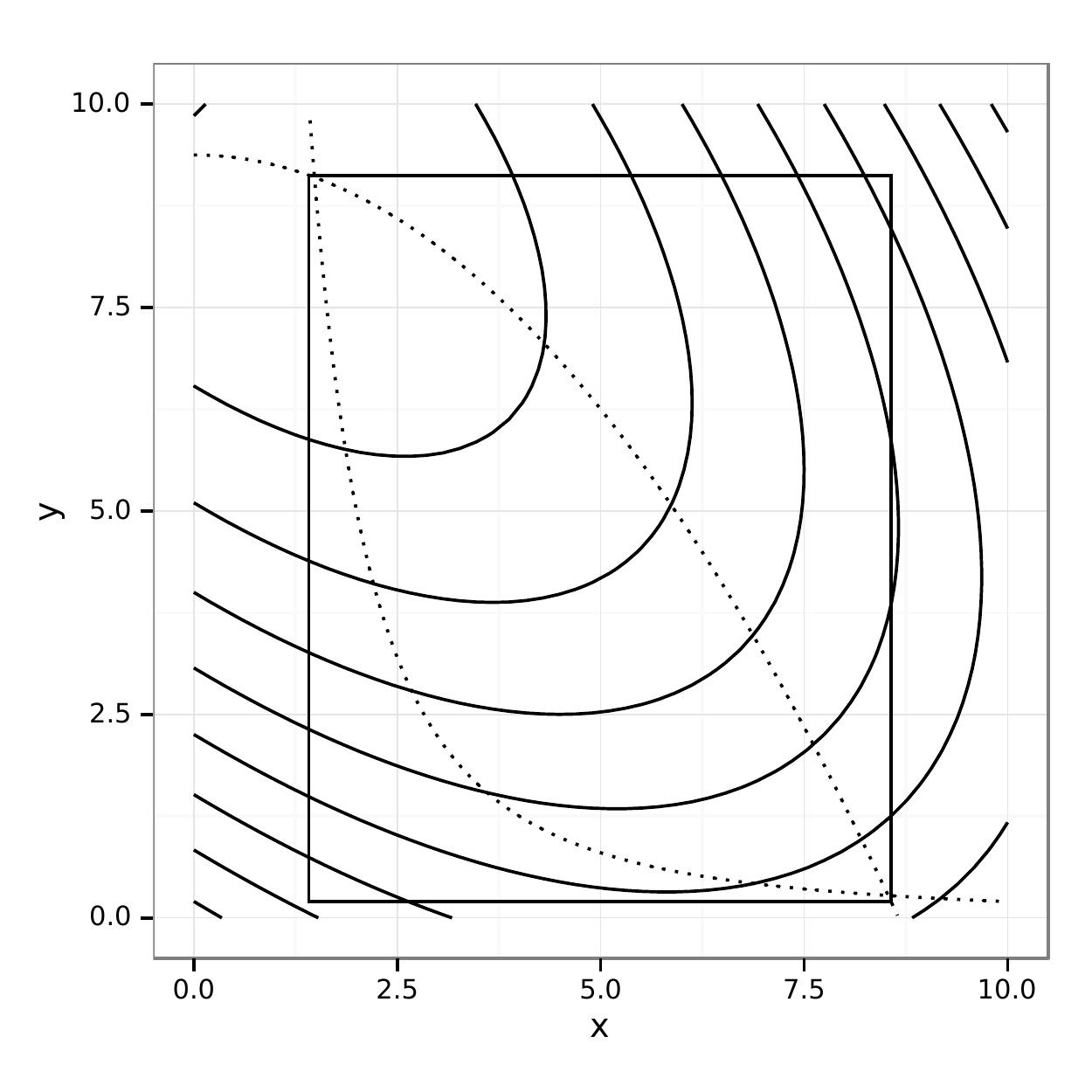}} \quad
	\subfloat[Convex hull at generation 10]{\label{fig:domain-10}
		\includegraphics[width=0.45\columnwidth]{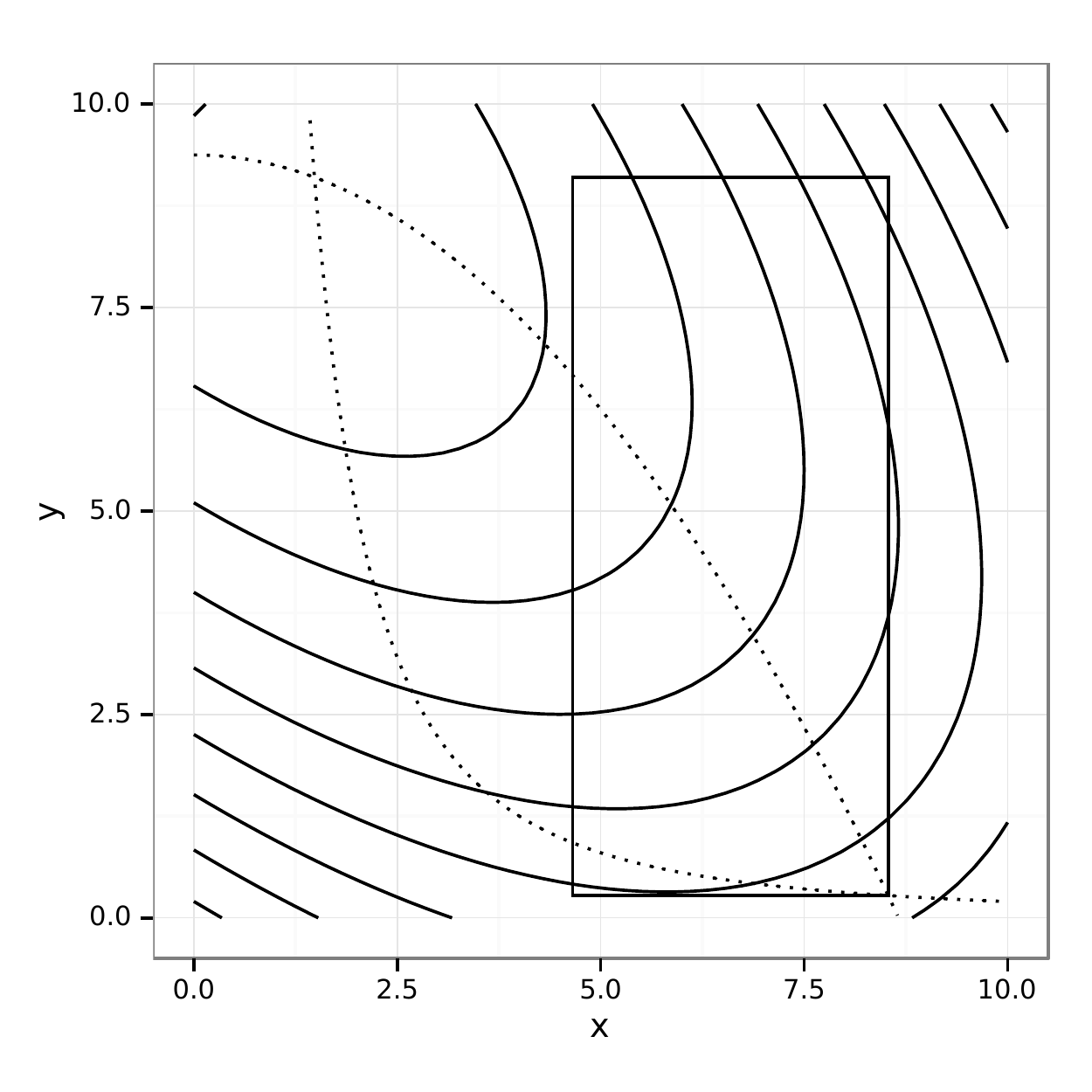}}
	\subfloat[Convex hull at generation 20]{\label{fig:domain-20}
		\includegraphics[width=0.45\columnwidth]{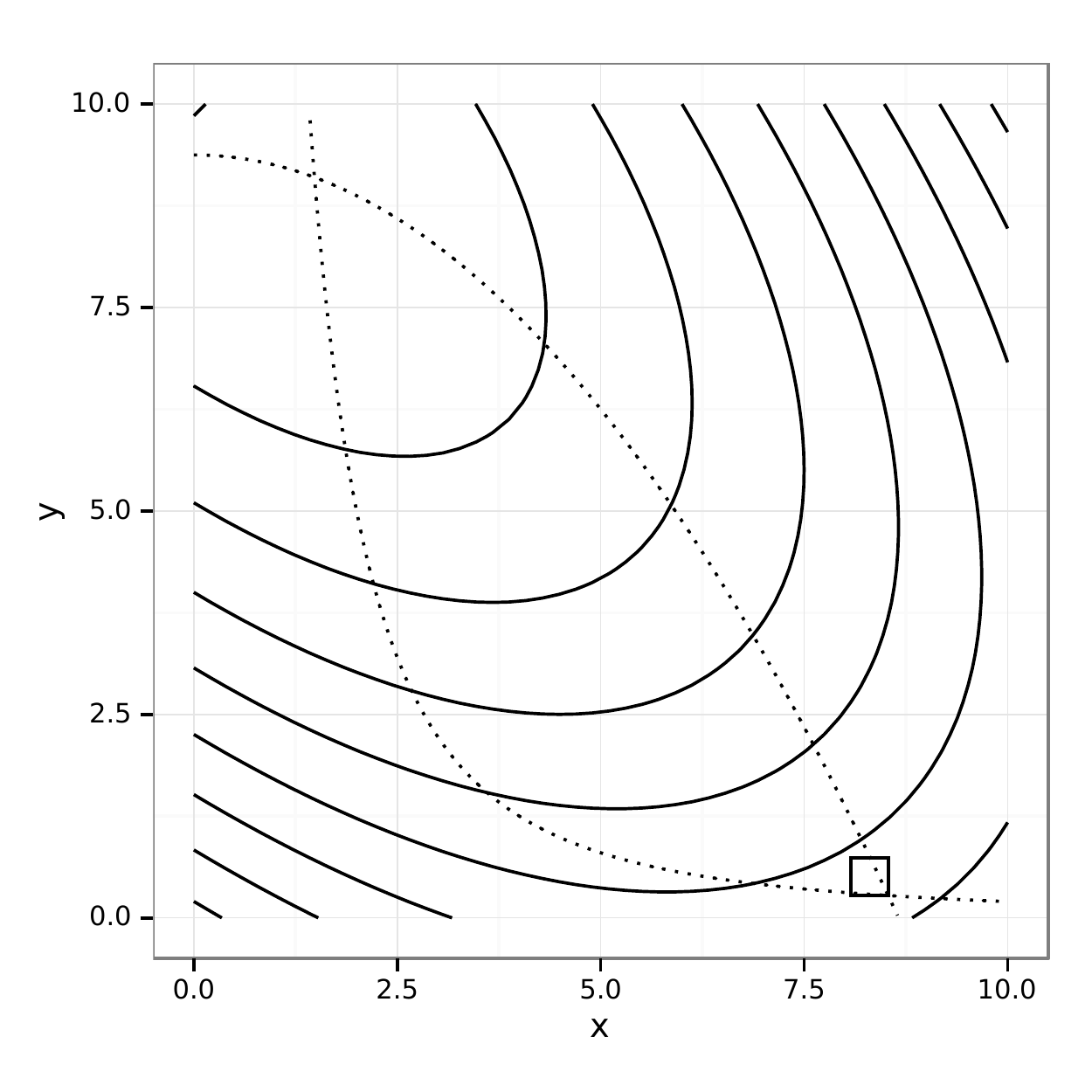}}
	\caption{Evolution of DE's domain with the number of generations}
\end{figure}

The following numerical example illustrates how the remaining boxes are exploited to reduce DE's
domain through the generations. Let
\begin{equation}
\begin{aligned}
\min_{(x, y) \in (X, Y)} \quad & -\frac{(x+y-10)^2}{30} - \frac{(x-y+10)^2}{120}  \\
\text{s.t.}	 \quad 	& \frac{20}{x^2} - y \le 0 \\
					& x^2 + 8y - 75 \le 0
\end{aligned}
\end{equation}
be a constrained optimization problem defined on the box $X \times Y = [0, 10] \times [0, 10]$
(Figure \ref{fig:initial-domain}). The dotted curves represent the frontiers of the two inequality
constraints, and the contour lines of the objective function are shown in solid dark. The feasible
region is the banana-shaped set, and the global minimizer is located in its lower right corner.

The initial domain of DE (which corresponds to the initial box in the IBC) is first contracted with
respect to the constraints of the problem. The initial population of DE is then generated within
this contracted domain, thus avoiding obvious infeasible regions of the search-space. This approach
is similar to that of~\cite{Focacci2003Local}. Figure \ref{fig:contracted-domain} depicts the
contraction (the black rectangle) of the initial domain with respect to the constraints
(sequentially handled by HC4Revise): $X \times Y = [1.4142, 8.5674] \times [0.2, 9.125]$.

Periodically, we compute the convex hull $\square(\mathcal{Q})$ of the remaining boxes of
$\mathcal{Q}$ and replace DE's domain with $\square(\mathcal{Q})$. Note that
\begin{enumerate}
\item the convex hull (linear complexity) may be computed at low cost, because the size
of $\mathcal{Q}$ remains small when using MaxDist;
\item by construction, MaxDist handles boxes on the rim of the remaining domain (the boxes of
$\mathcal{Q}$), which boosts the reduction of the convex hull.
\end{enumerate}
Figures \ref{fig:domain-10} and \ref{fig:domain-20} represent the \textit{convex hull}
$\square(\mathcal{Q})$ of the remaining subboxes in the IBC, respectively after 10 and 20 DE
generations. The population is then randomly regenerated within the new contracted domain
$\square(\mathcal{Q})$. The convex hull operation progressively eliminates local minima and
infeasible regions. The global minimum eventually found by Charibde with precision $\varepsilon =
10^{-8}$ is $\tilde{f} = f(8.532424, 0.274717) = -2.825296148$; both constraints are active.

\section{Experimental Results}\label{sec:results}

Currently, GlobSol~\cite{Kearfott1996Rigorous}, IBBA~\cite{Ninin2010Reliable}
and Ibex~\cite{ChabertJaulin2009} are among the most efficient solvers in rigorous constrained
optimization. They share a common skeleton of interval branch and bound algorithm, but differ in the
acceleration techniques. GlobSol uses the reformulation-linearization technique (RLT), that
introduces new auxiliary variables for each intermediary operation. IBBA calls a contractor similar
to HC4Revise, and computes a relaxation of the system of constraints using affine arithmetic.
Ibex is dedicated to both numerical CSPs and constrained optimization; it embeds most of the
aforementioned contractors (HC4, 3B, Mohc, CID, X-Newton).
Couenne~\cite{Belotti2009Branching} and BARON~\cite{Sahinidis1996BARON} are state-of-the-art NLP
solvers. They are based on a non-rigorous spatial branch and bound algorithm, in which the
objective function and the constraints are over- and underestimated by convex relaxations. 
Although they perform an exhaustive exploration of the search-space, they cannot guarantee a given
precision on the value of the optimum.

All five solvers and Charibde are compared on a subset of 11 COCONUT constrained problems (Table
\ref{tab:problem-descriptions}), extracted by Araya~\cite{Araya2012Contractor} for their difficulty:
ex2\_1\_7, ex2\_1\_9, ex6\_2\_6, ex6\_2\_8, ex6\_2\_9, ex6\_2\_11, ex6\_2\_12, ex7\_2\_3, ex7\_3\_5,
ex14\_1\_7 and ex14\_2\_7. Because of numerical instabilities of the ocaml-glpk LP library (``assert
failure''), the results of the problems ex6\_1\_1, ex6\_1\_3 and ex\_6\_2\_10 are not presented.
The second and third columns give respectively the number of variables $n$ and the number of
constraints $m$. The fourth (resp. fifth) column specifies the type of the objective function
(resp. the constraints): L is linear, Q is quadratic and NL is nonlinear. The logsize of the domain
$\bm{D}$ (sixth column) is $\log(\prod_{i=1}^n (\overline{D_i} - \underline{D_i}))$.

The comparison of CPU times (in seconds) for solvers GlobSol, IBBA, Ibex, Couenne, BARON and
Charibde on the benchmark of 11 problems is detailed in Table \ref{tab:comparison-solvers}.
Mean times and standard deviations (in brackets) are given for Charibde over 100 runs. The numerical
precision on the objective function $\varepsilon = 10^{-8}$ and the tolerance for equality
constaints $\varepsilon_= = 10^{-8}$ were identical for all solvers. TO (timeout)
indicates that a solver could not solve a problem within one hour.
The results of GlobSol (proprietary piece of software) were not available for all problems; only
those mentioned in~\cite{Ninin2010Reliable} are presented. The results of IBBA were also taken from
\cite{Ninin2010Reliable}. The results of Ibex were taken from \cite{Araya2012Contractor}: only the
best strategy (simplex, X-NewIter or X-Newton) for each benchmark problem is presented.
Couenne and BARON (only the commercial version of the code is available) were run on the NEOS
server~\cite{Gropp1997Optimization}.

% Couenne feas\_tolerance 1e-8
% BARON optca = 1e-8 optcr = 0

\setlength{\tabcolsep}{5pt}
\begin{table}[h!]
\centering
\caption{Description of difficult COCONUT problems}
\begin{tabular}{|l|cc|cc|c|}
\hline
			& 		&		& \multicolumn{2}{c|}{Type} & \\
Problem 	& $n$ 	& $m$ 	& $f$ 	& $g_i, h_j$ & Domain logsize \\
\hline
ex2\_1\_7	& 20	& 10	& Q & L & $+\infty$ \\
ex2\_1\_9	& 10	& 1		& Q & L & $+\infty$ \\
ex6\_2\_6	& 3		& 1		& NL & L & $-3 \cdot 10^{-6}$ \\
ex6\_2\_8	& 3		& 1		& NL & L & $-3 \cdot 10^{-6}$ \\
ex6\_2\_9	& 4		& 2		& NL & L & $-2.77$ \\
ex6\_2\_11	& 3		& 1		& NL & L & $-3 \cdot 10^{-6}$ \\
ex6\_2\_12	& 4		& 2		& NL & L & $-2.77$ \\
ex7\_2\_3	& 8		& 6		& L & NL & $61.90$ \\
ex7\_3\_5	& 13	& 15	& L & NL & $+\infty$ \\
ex14\_1\_7	& 10	& 17	& L	& NL & $23.03$ \\
ex14\_2\_7	& 6		& 9		& L & NL & $+\infty$ \\
\hline
\end{tabular}
\label{tab:problem-descriptions}
\end{table}

\setlength{\tabcolsep}{5pt}
\begin{table}[h!]
\centering
\caption{Comparison of convergence times (in seconds) between GlobSol, IBBA, Ibex, Charibde (mean
and standard deviation over 100 runs), Couenne and BARON on difficult constrained problems}
\begin{tabular}{|l|cccc|cc|}
\hline
			& \multicolumn{4}{c|}{Rigorous}	& \multicolumn{2}{c|}{Non rigorous} \\
Problem 	& GlobSol 	& IBBA	 	& Ibex 		& Charibde 		& Couenne & BARON \\
\hline
ex2\_1\_7	&			& 16.7		& \bf{7.74}	& 34.9 (13.3)			& 476 		& 16.23 \\
ex2\_1\_9	& 			& 154		& 9.07		& 35.9 (0.29)			& \bf{3.01}	& 3.58 \\
ex6\_2\_6	& 306		& 1575		& 136		& \textbf{3.3} (0.41) 	& TO		& 5.7 \\
ex6\_2\_8	& 204		& 458		& 59.3		& 2.9 (0.37)			& TO		& TO \\
ex6\_2\_9	& 463		& 523		& 25.2		& \textbf{2.7} (0.03)	& TO		& TO \\
ex6\_2\_11	& 273		& 140		& 7.51		& \textbf{1.96} (0.06)	& TO 		& TO \\
ex6\_2\_12	& 196		& 112		& 22.2		& \textbf{8.8} (0.17)	& TO		& TO \\
ex7\_2\_3	& 			& TO		& 544		& \textbf{1.9} (0.30)	& TO 		& TO \\
ex7\_3\_5	& 			& TO		& 28.91		& \textbf{4.5} (0.09) 	& TO		& 4.95 \\
ex14\_1\_7	& 			& TO		& 406		& 4.2 (0.13) 			& 13.86		& \bf{0.56} \\
ex14\_2\_7	& 			& TO		& 66.39		& 0.2 (0.04)			& \bf{0.01}	& 0.02 \\
\hline
Sum 		& $>$ 1442		& TO	& 1312.32	& \bf{101.26}			& TO		& TO \\
\hline
\end{tabular}
\label{tab:comparison-solvers}
\end{table}

Charibde was run on an Intel Xeon E31270 @ 3.40GHz x 8 with 7.8 GB of RAM. BARON and
Couenne were run on 2 Intel Xeon X5660 @ 2.8GHz x 12 with 64 GB of RAM. IBBA and Ibex were run on
similar processors (Intel x86, 3GHz). The difference in CPU time between computers is about
10\%~\cite{Araya2014Upper}, which makes the comparison quite fair.

The hyperparameters of Charibde for the benchmark problems are given in Table
\ref{tab:parametres-charibde}; $\mathit{NP}$ is the population size, and $\eta$ is the quasi-fixed 
point
ratio. The amplitude $W = 0.7$, the crossover rate $\mathit{CR} = 0.9$ and the MaxDist strategy are
common to all problems. Tuning the hyperparameters is generally problem-dependent, and requires
structural knowledge about the problem: the population size $\mathit{NP}$ may be set according
to the dimension and the number of local minima, the crossover rate $\mathit{CR}$ is related to the
separability of the problem, and the techniques based on linear relaxation have little influence
for problems with few constraints, but are cheap when the constraints are linear.

\begin{table}[h!]
	\centering
	\caption{Hyperparameters of Charibde for the benchmark problems}
	\begin{tabular}{|l|c|cccc|}
	\hline
	Problem 	& $\mathit{NP}$ 	& Bisections 	& Fixed-point ratio ($\eta$) 	& LP 		&
X-Newton \\
	\hline
	ex2\_1\_7	& 20				& RR			& 0.9		& \checkmark 	& \checkmark \\
	ex2\_1\_9	& 100				& RR			& 0.8		& \checkmark 	& \\
	ex6\_2\_6	& 30				& Smear			& 0			& \checkmark 	& \\
	ex6\_2\_8	& 30 				& Smear			& 0			& \checkmark 	& \\
	ex6\_2\_9	& 70				& Smear			& 0			& 				& \\
	ex6\_2\_11	& 35				& Smear			& 0			& 				& \\
	ex6\_2\_12	& 35				& RR			& 0			& \checkmark 	& \\
	ex7\_2\_3	& 40				& Largest		& 0			& \checkmark 	& \checkmark \\
	ex7\_3\_5	& 30				& RR			& 0			& \checkmark 	& \\
	ex14\_1\_7	& 40				& RR			& 0			& \checkmark 	& \\
	ex14\_2\_7	& 40				& RR			& 0			& \checkmark 	& \\
	\hline
	\end{tabular}
	\label{tab:parametres-charibde}
\end{table}

Charibde outperforms Ibex on 9 out of 11 problems, IBBA on 10 out of 11 problems
and GlobSol on all the available problems. The cumulated CPU time shows that Charibde (101.26s)
improves
the performances of Ibex (1312.32s) by an order of magnitude (ratio: 13) on this benchmark of 11
difficult problems. Charibde also proves highly competitive against \textit{non-rigorous} solvers
Couenne and BARON. The latter are faster or have similar CPU times on some of the 11 problems,
however they both timeout on at least five problems (seven for Couenne, five for BARON). Overall,
Charibde seems more robust and solves all the problems of the benchmark, while providing a numerical
proof of optimality. Surprisingly, the convergence times do not seem directly related to the
dimensions of
the instances. They may be explained by the nature of the objective function and constraints
(in particular, Charibde seems to struggle when the objective function is quadratic) and the
dependency induced by the multiple occurrences of the variables.

Table \ref{tab:upper-bounds} presents the best upper bounds obtained by Charibde, Couenne and BARON
at the end of convergence (precision reached or timeout). Truncated digits on the upper bounds
are bounded (e.g. $1.23_7^8$ denotes $[1.237, 1.238]$ and $-1.23_7^8$ denotes $[-1.238, -1.237]$).
The incorrect digits of the global minima obtained by Couenne and BARON are underlined. This
demonstrates that non-rigorous solvers may be affected by roundoff errors, and may provide solutions
that are infeasible or have an objective value \textit{lower than the global minimum} (Couenne
on ex2\_1\_9, BARON on ex2\_1\_7, ex2\_1\_9, ex6\_2\_8, ex6\_2\_12, ex7\_2\_3 and ex7\_3\_5). For
the most difficult instance ex7\_2\_3, Couenne is not capable of finding a feasible solution with a
satisfactory evaluation within one hour. It would be very informative to compute the ratio between
the size of the feasible domain (the set of all feasible points) and the size of the entire domain.
On the other hand, the strategy MaxDist within Charibde
greatly contributes to finding an excellent upper bound of the global minimum, which significantly
accelerates the interval pruning phase.

\begin{table}[h!]
	\centering
	\caption{Best upper bounds obtained by Charibde, Couenne and BARON}
	\begin{tabular}{|l|ccc|}
	\hline
	Problem 	& Charibde 				& Couenne 				& BARON\\
	\hline
	ex2\_1\_7	& $-4150.41013392_8^9$ 	& $-4150.4101\ul{2731}_7^8$	& $-4150.4101\ul{6079}_7^8$ \\%
	ex2\_1\_9	& $-0.3750000075$ 		& $-0.375000\ul{15}_3^4$ 	& $-0.37500\ul{111}_0^1$ \\%
	ex6\_2\_6	& $-0.00000260_2^3$ 	& $0.00000071_0^1$ 		& $-0.00000260_2^3$  \\%
	ex6\_2\_8	& $-0.0270063_{49}^{50}$	& $-0.0270063_{49}^{50}$ & $-0.0270063\ul{7}_0^1$ \\%
	ex6\_2\_9	& $-0.03406618_4^5$ 	& $-0.034066184$ 		& $-0.03406619_0^1$ \\
	ex6\_2\_11	& $-0.00000267_2^3$ 	& $-0.00000267_2^3$		& $-0.00000267_2^3$ \\%
	ex6\_2\_12	& $0.2891947_{39}^{40}$	& $0.28919475$ 			& $0.28919\ul{169}_8^9$ \\%
	ex7\_2\_3	& $7049.24802052_8^9$ 	& $10^{50}$ 			& $7049.\ul{02029170}_6^7$ \\%
	ex7\_3\_5	& $1.20671699_1^2$ 		& $1.2068965$ 			& $\ul{0.23982448}_7^8$ \\%
	ex14\_1\_7	& $0.0000000_{09}^{10}$	& $0.00000000_0^1$ 		& $0$ \\%
	ex14\_2\_7	& $0.00000000_7^8$ 		& $0.00000000_0^1$ 		& $0$ \\%
	\hline
	\end{tabular}
	\label{tab:upper-bounds}
\end{table}

\section{Conclusion}
We proposed a cooperative hybrid solver Charibde, in which a deterministic
interval branch and contract cooperates with a stochastic differential evolution algorithm. The two
independent algorithms run in parallel and exchange bounds, solutions and search-space in an
advanced manner via message passing. The domain of the population-based metaheuristic is
periodically reduced by removing local minima and infeasible regions detected by the branch and
bound.

A comparison of Charibde with state-of-the-art interval-based solvers (GlobSol, IBBA, Ibex) and NLP
solvers (Couenne, BARON) on a benchmark of difficult COCONUT problems shows that Charibde is highly
competitive against non-rigorous solvers (while bounding the global minimum) and converges faster
than rigorous solvers by an order of magnitude.

\clearpage

\bibliographystyle{splncs03}
\bibliography{Vanaret}
\end{document}

%% file: de.pdf_tex
%% Creator: Inkscape inkscape 0.48.3.1, www.inkscape.org
%% Accompanies image file 'de.pdf' (pdf, eps, ps)
\begingroup%
  \makeatletter%
  \providecommand\color[2][]{%
    \errmessage{(Inkscape) Color is used for the text in Inkscape, but the package 'color.sty' is not loaded}%
    \renewcommand\color[2][]{}%
  }%
  \providecommand\transparent[1]{%
    \errmessage{(Inkscape) Transparency is used (non-zero) for the text in Inkscape, but the package 'transparent.sty' is not loaded}%
    \renewcommand\transparent[1]{}%
  }%
  \providecommand\rotatebox[2]{#2}%
  \ifx\svgwidth\undefined%
    \setlength{\unitlength}{228.8bp}%
    \ifx\svgscale\undefined%
      \relax%
    \else%
      \setlength{\unitlength}{\unitlength * \real{\svgscale}}%
    \fi%
  \else%
    \setlength{\unitlength}{\svgwidth}%
  \fi%
  \global\let\svgwidth\undefined%
  \global\let\svgscale\undefined%
  \makeatother%
  \begin{picture}(1,0.66433566)%
    \put(0,0){\includegraphics[width=\unitlength]{de.pdf}}%
    \put(0.7964167,0.31588686){\color[rgb]{0,0,0}\makebox(0,0)[lb]{\smash{minimum}}}%
    \put(0.2232447,0.54685099){\color[rgb]{0,0,0}\makebox(0,0)[lb]{\smash{$\bm{w}$}}}%
    \put(0.39450036,0.39498661){\color[rgb]{0,0,0}\makebox(0,0)[lb]{\smash{$\bm{v}$}}}%
    \put(0.49360443,0.22537928){\color[rgb]{0,0,0}\makebox(0,0)[lb]{\smash{$\bm{u}$}}}%
    \put(0.62777417,0.1715122){\color[rgb]{0,0,0}\makebox(0,0)[lb]{\smash{$\bm{y}$}}}%
    \put(0.92748319,0.05487541){\color[rgb]{0,0,0}\makebox(0,0)[lb]{\smash{$x_1$}}}%
    \put(0.05088425,0.61921033){\color[rgb]{0,0,0}\makebox(0,0)[lb]{\smash{$x_2$}}}%
    \put(0.66823962,0.49499229){\color[rgb]{0,0,0}\makebox(0,0)[lb]{\smash{$W \times (\bm{v}-\bm{w})$}}}%
    \put(0.4540696,0.61300253){\color[rgb]{0,0,0}\makebox(0,0)[lb]{\smash{$\bm{v}-\bm{w}$}}}%
    \put(0.17281819,0.18597199){\color[rgb]{0,0,0}\makebox(0,0)[lb]{\smash{$f$}}}%
    \put(0.37333922,0.16176008){\color[rgb]{0,0,0}\makebox(0,0)[lb]{\smash{$\bm{x}$}}}%
  \end{picture}%
\endgroup%

%% file: charibde.pdf_tex
%% Creator: Inkscape inkscape 0.48.3.1, www.inkscape.org
%% Accompanies image file 'charibde.pdf' (pdf, eps, ps)
\begingroup%
  \makeatletter%
  \providecommand\color[2][]{%
    \errmessage{(Inkscape) Color is used for the text in Inkscape, but the package 'color.sty' is
not loaded}%
    \renewcommand\color[2][]{}%
  }%
  \providecommand\transparent[1]{%
    \errmessage{(Inkscape) Transparency is used (non-zero) for the text in Inkscape, but the package
'transparent.sty' is not loaded}%
    \renewcommand\transparent[1]{}%
  }%
  \providecommand\rotatebox[2]{#2}%
  \ifx\svgwidth\undefined%
    \setlength{\unitlength}{468.4bp}%
    \ifx\svgscale\undefined%
      \relax%
    \else%
      \setlength{\unitlength}{\unitlength * \real{\svgscale}}%
    \fi%
  \else%
    \setlength{\unitlength}{\svgwidth}%
  \fi%
  \global\let\svgwidth\undefined%
  \global\let\svgscale\undefined%
  \makeatother%
  \begin{picture}(1,0.34599488)%
    \put(0,0){\includegraphics[width=\unitlength]{charibde.pdf}}%
   
\put(0.05685006,0.32103146){\color[rgb]{0,0,0}\makebox(0,0)[lt]{\begin{minipage}{
0.2208125\unitlength}\centering Differential Evolution\end{minipage}}}%
    \put(0.11708794,0.11658555){\color[rgb]{0,0,0}\makebox(0,0)[lb]{\smash{population}}}%
    \put(0.05485907,0.19819186){\color[rgb]{0,0,0}\makebox(0,0)[lb]{\smash{best individual}}}%
    \put(0.70452608,0.19819186){\color[rgb]{0,0,0}\makebox(0,0)[lb]{\smash{best upper bound}}}%
    \put(0.70602293,0.11504706){\color[rgb]{0,0,0}\makebox(0,0)[lb]{\smash{punctual solution}}}%
   
\put(0.67907182,0.32222638){\color[rgb]{0,0,0}\makebox(0,0)[lt]{\begin{minipage}{
0.29850955\unitlength}\centering Interval Branch and Contract\end{minipage}}}%

    \put(0.70736555,0.03920418){\color[rgb]{0,0,0}\makebox(0,0)[lb]{\smash{subspaces}}}%
    \put(0.17111336,0.03999746){\color[rgb]{0,0,0}\makebox(0,0)[lb]{\smash{domain}}}%
    \put(0.4373326,0.22383309){\color[rgb]{0,0,0}\makebox(0,0)[lb]{\smash{updates}}}%
    \put(0.40746615,0.087837){\color[rgb]{0,0,0}\makebox(0,0)[lb]{\smash{injected into}}}%
    \put(0.44308528,0.01341954){\color[rgb]{0,0,0}\makebox(0,0)[lb]{\smash{reduce}}}%
  \end{picture}%
\endgroup%